# Solving Sinhala Language Arithmetic Problems using Neural Networks

*ᵗ* Chathurika W M T [1], *ᵗ* De Silva K C E [2], *ᵗ* Raddella A M [3], *ᵗ* Ekanayake E M R S [4],

*Nugaliyadde A [5], Mallawarachchi Y [6]

Sri Lanka Institute of Information Technology, Sri Lanka
{it12142538[1], it12143214[2], it12082148[3],
it12142774[4]}@my.sliit.lk, {anupiya.n[5], yashas.m[6]}@sliit.lk

*ABSTRACT*

*A methodology is presented to solve Arithmetic problems in Sinhala Language using a Neural Network. The system comprises of (a) keyword identification, (b) question identification, (c) mathematical operation identification and is combined using a neural network. Naïve Bayes Classification is used in order to identify keywords and Conditional Random Field to identify the question and the operation which should be performed on the identified keywords to achieve the expected result. "One vs. all Classification" is done using a neural network for sentences. All functions are combined through the neural network which builds an equation to solve the problem. The paper compares each methodology in ARIS and Mahoshadha to the method presented in the paper. Mahoshadha2 learns to solve arithmetic problems with the accuracy of 76%.*

## 1. INTRODUCTION

Natural Language Processing (NLP) has become one of the most prominent research areas with the development of Artificial Neural Networks (ANN). Question Answering (QA) Systems have been highly worked on research area in NLP. IBM Watson [1] drew attention for QA systems from and increased the work on them. QA systems can be pointed in two different avenues, Corpus-based QA systems and Natural Language Mathematics based QA systems.

Corpus-based QA systems rely on the data which is available to the system. The Question will be broken down and keywords will be identified in order to traverse through the available corpus to and find the relevant information. A Language-based Mathematical QA system will focus will on not only the keywords but the actions in the performed in the question. It will focus on the addition and subtraction on a focused quantity.

Although there are many QA systems available, they are language specific. Sinhala QA systems have been given a very less focus. The recent publication of "Mahoshadha" [6] showed promise in QA Systems for the Sinhala Language.

*ᵗ Equally Contributed in Developing the Research  *Designed & Conceptualized the Research*

This paper has improved on "Mahoshadha" adding it the ability to understand Mathematical problems given in the Sinhala Language. This focuses on identifying the relevant keywords using Conditional Random Fields (CRF) and joining them together with an ANN.

In section 1 related work is described, section 2 describes the methodology, and section 3 describes the results. The paper shows an in-depth view of approaching Mathematical QA system for the Sinhala Language.

## 2. RELATED WORK

There are prominent researches which are based on QA on IQ questions. It was found that the task of a question answering system [1, 12] is finding the exact answer to a given question by finding out through the relevant documentation on the internet. Most of these systems are not working on solving mathematical sentence problems.

The focus of research has been on understanding the semantics of a natural language text. There exist QA Systems such as 'IBM Watson' [1], 'Mahoshadha' [6] and math-aware systems like Math Find and ARIS [5].

"Mahoshadha" [6] was developed using a standard annotated Sinhala Corpus. The research was capable of answering questions from the available corpus. The content will be extracted from the corpus in order to answer the questions. Since it is a corpus-based QA system it is not capable of answering Arithmetic problems given.

"ARIS" [5] is a mathematical problem-solving system for the English language. The system is based on variable / value analyzing, keyword extraction. ARIS [5] only relies on learning verb categories which alleviate the need for equation templates for arithmetic problems in English. The proposed system distinguishes from this system by using all above methods in the Sinhala language.

Recent work are more focused on content extraction in which the system search through unstructured data in machine readable format to find out a solution for a given problem. In our endeavor, the main intention was to make the system enable to solve a particular problem by itself. Most of the existing systems do 'WordNet' based similarity measures that typically depend heavily on IS-A information, which is available for nouns but incomplete for verbs and completely lacking for adjectives and adverbs [12].

## 3. METHODOLOGY

The methodology describes a developed platform to analyze and understand an arithmetic problem in the Sinhala language. The functionality provided is intended to guide the user to work efficiently and reduce the time consumption to solve arithmetic problems. The system is a combination of 3 components combined with an ANN. The components are Keyword Identification, Question Identification, and Arithmetic Operation Identification.

Parts of Speech (POS) tagging process is used to identify nouns, verbs and etc. of any sentence. With the use of POS tagging, the task will provide an opportunity to identify important keywords from a given question.

The proposed system selects out nouns, verbs, and adjectives from a question. In order to achieve POS tagging, a training dataset was used. Sinhala Mathematical textbooks were used in order to achieve this task.

The questions will be divided by the sentences by the full stops and question marks. The words in the sentence are divided by the spaces. The identified words will be passed to the next function which identifies the word categories.

Two kinds of ambiguous situations that arise mainly in Sinhala are handled by the model designed in the work. Known Word Ambiguity Words often belong to two or more syntactic categories (can have two or more POS Tags). For an example

අඹගෙඩි_NN

අඹ_ADJ

ගෙඩි_NN

Even though the same word may belong to several categories system can sort out the most suitable tag for the word by since the word has occurred in several contexts in the training data set.

Word Ambiguity occurs when word that does not exist in the train data set is given. In such a situation system will compare the word with train data to find out the most similar word and tag accordingly. Most of the available classifications have been made with the intention of teaching/learning the language. Since this kind of classification is not suitable for an arithmetic problem-solving system. The system developed a more comprehensive classification for the purpose of defining POS tags. Given an arithmetic problem system will tag the words as given in the example below.

මම ළෙ අඹගෙඩි 10 ක් ඇත.

මම_PRO ළෙ_PP අඹගෙඩි_NN 10_CD ක්_PP ඇත_VB.

A word to be classified may be a noun, verb, number or adjective and etc. By using naïve Bayes classifier, the system is trained to identify the relevant tag of the word by identifying the correct word category it falls into.

To experimentally evaluate the method a dataset of arithmetic word problems along with the correct solutions. The method was tested on the accuracy of solving arithmetic word problems and identifying each tag categories in sentences.

Question Identification process Conditional Rational Fields tool is used (CRF++ tool).Conditional Random Fields (CRFs) are a probabilistic framework for labeling and segmenting structured data, such as sequences and trees [18]. It is capturing known relationships between observations and constructs consistent interpretations. CRFs outperform both Maximum Entropy Markov Models and HMMs on a number of real-world tasks in many fields, including bioinformatics, computational linguistics and speech recognition. The main advantage of CRFs over Hidden Markov Models (HMM) is their conditional nature. In a CRF, each feature is a function that takes in as input. One possible feature function could predict the label of the current word given that the previous word label. Afterward, CRF assigns a weight for each feature function and later on convert that score into a probabilistic value in-between 0 and 1.

$$score(l|s) = \sum_{j=1}^{m} \sum_{i=1}^{n} \lambda_j f_j(s, i, l_i, l_{i-1})$$

For a given sentence '**s**', by adding up the weighted features over all words in the sentence a score is given to a labeling **l** of **s** as above.

$$p(l|s) = \frac{exp[score(l|s)]}{\sum_{l'} exp[score(l'|s)]} = \frac{exp[\sum_{j=1}^{m} \sum_{i=1}^{n} \lambda_j f_j(s,i,l_i,l_{i-1})]}{\sum_{l'} exp[\sum_{j=1}^{m} \sum_{i=1}^{n} \lambda_j f_j(s,i,l'_i,l'_{i-1})]}$$

Here the scores are getting transformed into probabilities **p(l|s)** between 0 and 1 by exponentiation and normalizing.

Example of a feature function that could be used as:-

$$f1(s, i, li, li-1) = 1 \text{ if } li = NEP \text{ ; } 0 \text{ otherwise}$$

In Question identification Feature Selection, Generation of Feature Template and Train and Test the CRF Algorithm takes place. In any statistical model, feature functions act the critical role. ML algorithm learns patterns of the data set using these features. The features which have been used by Hindi, Bengali and Arabic NER systems were selected. Since Sinhala is an Indic language, features used by South and Southeast Asian languages are more applicable. Feature template is a file required by CRF++ tool. Feature template describes which features are used in training and testing. Each line in the template file denotes one template. In each template, special macro %x [row, column] will be used to specify a token in the input data. Row specifies the relative position from the current focusing token and column specifies the absolute position of the column. All the feature columns were fed into CRF++ via this template file.

When training CRF++ toolkit which is using CRF data-driven technique has been used for the research project. It requires two separate training and testing data sets. CRF++ strictly requires both the training and testing datasets to be in the same format.

The mathematical operation to be done on the question is identified to calculate the answer. Solving a problem consists of two main steps: (1) Progressing states based on verb categories in sentences and (2) Finding the mathematical operation. The research mainly focuses on solving addition and subtraction problems.

Once the system finds the consisting of entities (E), Containers (C), Attributes (A), relations(R) among entities and quantities (take as the inputs from keyword identification and mathematical question identification) input the tagged sentences one by one. Then it forms the states $< s_0, s_1, s_2 \ldots s_T >$. After forming states identify the verb category correctly. There are seven verb categories.

System match the entities and categories (An entity/category is matched if has the same headword and the same set of attributes as an existing entity/category). If an entity or category cannot be matching, then a new one is created. For the matched entities, initializes or updates the values of the containers in the state. After updating the values in the containers in the states which are matched find the relevant mathematical operation and forms the equation by comparing the quantities of matched containers between two states.

The ANN categorizes the mathematical problem sentence into two categories as 'Positive' and 'Negative' using "One vs. All Classification" ANN. Only two categories are taken into consideration a binary classification takes place in this section. According to the sentence features explained in the train data set, ANN categorizes the sentences under the positive category. Once the categorization is done the other set of sentences which were not categorized are taken into consideration. Those are classified under the negative category.

In order to perform that, the system maintains a dataset of different types and forms of the sentence and question patterns in the Sinhala language. This feature is the main key to lead the system to recognize the way of solving a given mathematical problem. This is a classification ANN in which the system identify the positive and negative nature of the sentences in a given mathematical word problem.

මා ළඟ අඹ ගෙඩි 10 ක් ඇත . 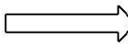 **POSITIVE**

ඉන් 2 ක් මල්ලිට දුන් විට තව ග ෙපමණ ඉතුරුද? 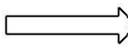 **NEGATIVE**

The intention of doing that is to separately identify the negative numeric values and the positive numeric values of a mathematical problem.

මා ළඟ අඹ ගෙඩි 10 ක් ඇත .                     ඉන් 2 ක් මල්ලිට දුන් විට තව ග ෙපමණ ඉතුරුද?

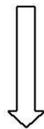     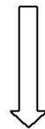

**POSITIVE**                    **NEGATIVE**

The above example, the system identifies the positive and negative nature of a sentence and classifies the sentence into positive or negative, then the numeric value.

Each layer of the ANN comprises of keyword identification, question identification, mathematical operation identification and classification together to form a complete system. In order to perform that, the outputs of the keyword identification, question identification, and mathematical operation identification are taken as inputs to the ANN in order to make a combination of subsystems.

Future improvements of making system able of performing operations like multiplication, division; derivation and etc., the multi-class classification must be implemented. The advantage of using "One vs. All Classification" is that the future improvements of the system are easy to be implemented.

## 4. RESULTS

The testing and training corpuses were taken from the Grade 5 student textbooks published by education publication department in 2010. Total of 200 Question were tested. Non-Language Arithmetic problems were eliminated. Non-straightforward word problems and other arithmetic problems could be solved with Mahoshadha2.

Keyword identification experimentally evaluates a method and builds a dataset of arithmetic word problems. The method is tested on the accuracy of solving arithmetic word problems and identifying each tag categories in sentences. A special corpus was used in order to identify distinct words.

Sentence categorization is performed with high accuracy even if the problem types and also the verbs are different. This is performed through a "One vs. All Classification" ANN. A training dataset of possible question forms is maintained. The classification in this system is a form of a binary classification. In order to test the feature, single sentences were given as test data. Then the ANN was tested with complete arithmetic problems presented in natural Sinhala language. The Same set of questions that were tested under keyword identification was given to the system for testing.

Mahoshadha 2 learns to classify sentences and does not require observing similar patterns/templates in the training data. Therefore, Mahoshadha 2 is more robust to differences between the training and test datasets and can generalize across different dataset types.

The datasets that was tested had mathematically similar problems but differ in the natural language properties such as in the sentence length and irrelevant information. The tests were done for problems based on addition and subtraction. The study generalizes a certain question pattern according to the train data sets.

Some of the sample questions that the system was tested for are as follows

1. "වට්ටියක සුදු මල් 12 ක් ඇත . රතු මල් 10 ක් ඇත . තවත් සුදු මල් 3 ක් එයට දැමුවා . සුදු මල් ගණන කොපමණද ?"
2. "බස්නැවතුම්පොලක සිටින ගැහැණු ගණන 10 කි . පිරිමි ගණන 20 කි . තවත් බසයකින් පැමිණ ගැහැණු 20 ක් බැස්සේය . පිරිමි 12 ක් බැස්සේය . බස්නැවතුම්පොලේ සිටින මුළු පිරිමි ගණන කොපමණද ?"
3. "ටැංකියක මාළු 10 ක් සිටි . එයට තවත් මාළු 4 ක් දැමුවිට මුළු මාළු ගණන කොපමණද ?"
4. ගෙවත්තේ අඹ ගස් 2 ක් තිබේ . ජම්බු ගස් 3 ක් ද තිබේ . සහ කොස් ගස් 5 ක් තිබේ. ගෙවත්තේ ඇති මුලු ගස් ගණන කියද ?
5. පාපැදියක් රුපියල් 8500 විය . රුපියල් 4800 ක් මදි විය . ඉතිරි මුදල මිතුරෙකුගෙන් ලබාගත්තෙමි. මා ලබාගත් මුදල කියද?

    Consider the second example question given above. A breakdown of the question

done by the system is given below.

බස්නැවතුම්පොලේ සිටින මුළු පිරිමි ගණන කොපමණද

⇩

System identifies the adjective under 'question identification'

බස්නැවතුම්පොලක සිටින ගැහැණු ගණන 10 කි
පිරිමි ගණන 20 කි **POSITIVE**
තවත් බසයකින් පැමිණ ගැහැණු 20 ක් බැස්සේය
පිරිමි 12 ක් බැස්සේය **NEGATIVE**

Only the sentences with the identified adjective are classified

Under the 'Mathematical operation identification' the mathematical operation is identified as 'addition'.

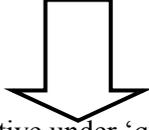

**Final answer is 32**

The 100 training datasets with high variations are used to train the ANN. The optimal dataset size was defined in order to increase the accuracy of the results. There are 3 separate datasets that have been used for 'keyword identification', 'Question identification' and 'classification neural network'. Certain words in Sinhala language such as 'ළඟ' 'ලඟ' should be included in the same manner in all 3 data sets. (if training data of neural network does not contain the word 'ළඟ' the system would not generate the correct output even though other datasets include it ).

The system was tested for the overall functionality in the similar way that the system was tested for keyword identification and classification. The same set of questions was tested first to maintain a flow throughout the testing process. The accuracy level of 74% in keyword identification and 72% success rate of sentence classification.

For the data set of 100 questions presented in the Sinhala language that was tested, correct solution was generated for 76 questions.

With the results gained, it can be stated that the system approximately has 76% of accuracy.

## 5. DISCUSSION

"Mahoshadha 2" is an extension of "Mahoshadha" [6] which is an intelligent QA system for the Sinhala language. This does not solve arithmetic problems. Moving a step forward from it the paper describes a novel approach to solving Sinhala arithmetic problems. The research only focuses on keywords and traversing through the corpus to find the answer from the keyword set.

ARIS [5] research is based on verb categorization and named entity recognition, by mapping the verbs in the mathematical problem into categories that describe the impact of verbs on world state. It identifies the entities with numeric values bound to those entities. Then build a mathematical equation in order to solve it. The approach in "Mahoshadha 2" the verb categorization takes place under the mathematical operation identification. This is not dependent on language features when handling a morphologically rich language like "Sinhala". The verb categorization is easy in a language like English but in "Sinhala" there are various forms of a verb which generate distinct meanings which are different.

ARIS research paper, the word problems where two entities exist, the second entity is identified as an attribute of the first entity by the ARIS system. For an example in a problem like

"There are 5 apples and 10 oranges in a basket. Two oranges are being eaten. How many oranges are there in the basket now?"

ARIS will identify oranges as an attribute of apples.

The presented system will identify that the system is expected to find a number of oranges, using CRF+ tool.

The most significant difference in between ARIS and Mahoshadha 2 is the manipulation of ANN. According to ARIS, generating the solution consists of two main steps as (1) progressing states based on verb categories in sentences and (2) forming the equation.

Mahoshadha 2 distributes the operations as keyword identification, question identification, and mathematical operation identification which will be layers of the ANN which at the same time generates the final output.

This paper is one step toward building a system that can solve any arithmetic problems given in Sinhala. The research only focuses on 'addition' and 'subtraction'. This can be extended to solve multiplication and division based problems since the baselines of other mathematical operations are addition and subtraction.

## 6. FUTURE WORK

The system's current capability can be extended for the task of multiplication and division by using One Vs All Classification in categorizing the problem in to four categories depending on the functionality required for calculation. There is also a probability in increasing the accuracy and allowing further accurate classification by introducing a Deep Neural Network. Deep Neural Networks have shown high performance in the past years in classification and QA systems. The Deep Neural Network can also support by removing the other algorithms which are currently used in presented research and allow the QA system to be solely based on a Deep Neural Network.

## 7. CONCLUSION

This paper describes an approach to solving arithmetic problems given in Sinhala Language using an ANN. The ANN is trained using the Sinhala grade 5 student textbooks published by education publication department in 2010. The efficiency of this system has been further enhanced by using POS filtering in keyword identification. In Question identification, the significance of the words in the problem is learned by the system. The relevant mathematical operation is identified by the verb categorization process. One-vs.-All classification ANN categorizes the sentences of a problem as 'Positive' and 'Negative' in the approach to solving the problem and then aggregates the above functions into a single system. The system was able to generate 76% of accurate results.